\pdfoutput=1
\documentclass[11pt]{article}
\usepackage{booktabs}

\usepackage[final]{acl}

\usepackage{times}
\usepackage{latexsym}

\usepackage[T1]{fontenc}
\usepackage[utf8]{inputenc}

\usepackage{microtype}

\usepackage{inconsolata}

\usepackage{graphicx}
\usepackage{CJKutf8}
\usepackage{multirow}
\usepackage{bbding}
\usepackage{xcolor}
\usepackage{diagbox}
\usepackage{makecell}
\usepackage{amsmath}
\usepackage{float}

\usepackage{color,xcolor}

\title{MIRAGE: Exploring How Large Language Models Perform in Complex Social Interactive Environments}

\author{
    \textbf{Yin Cai\textsuperscript{1}},
    \textbf{Zhouhong Gu\textsuperscript2},
    \textbf{Zhaohan Du\textsuperscript{2}},
    \textbf{Zheyu Ye\textsuperscript{4}},
    \\
    \textbf{Shaosheng Cao\textsuperscript{4}},
    \textbf{Yiqian Xu\textsuperscript{3}},
    \textbf{Hongwei Feng\textsuperscript{2*}},
    \textbf{Ping Chen\textsuperscript{1*}}
    \\
    \textsuperscript{1}Institute of Big Data, Fudan University\\
    \textsuperscript{2}Shanghai Key Laboratory of Data Science, School of Computer Science, Fudan University\\
    \textsuperscript{3}School of Computer Science, Fudan University, 
    \textsuperscript{4}Xiaohongshu Inc.
    \\
    \small{\{ycai25, zhgu22, duzh22\}@m.fudan.edu.cn, \{xuyiqian, hwfeng, pchen\}@fudan.edu.cn}\\
    \small{\{tanghuang1, guoba1\}@xiaohongshu.com}
}
\begin{document}
\begin{CJK}{UTF8}{gbsn}
\maketitle
\begin{abstract}
Large Language Models (LLMs) have shown remarkable capabilities in environmental perception, reasoning-based decision-making, and simulating complex human behaviors, particularly in interactive role-playing contexts. This paper introduces the Multiverse Interactive Role-play Ability General Evaluation (MIRAGE), a comprehensive framework designed to assess LLMs' proficiency in portraying advanced human behaviors through murder mystery games. MIRAGE features eight intricately crafted scripts encompassing diverse themes and styles, providing a rich simulation. To evaluate LLMs' performance, MIRAGE employs four distinct methods: the Trust Inclination Index (TII) to measure dynamics of trust and suspicion, the Clue Investigation Capability (CIC) to measure LLMs' capability of conducting information, the Interactivity Capability Index (ICI) to assess role-playing capabilities and the Script Compliance Index (SCI) to assess LLMs' capability of understanding and following instructions. Our experiments indicate that even popular models like GPT-4 face significant challenges in navigating the complexities presented by the MIRAGE. The datasets and simulation codes are available in \href{https://github.com/lime728/MIRAGE}{github}.
\end{abstract}

\section{Introduction}
\footnotetext[1]{Corresponding Authors}

Large Language Models (LLMs) have demonstrated remarkable potential in environmental perception and reasoning-based decision-making~\cite{xi2023rise, guo2024large, gu2024xiezhi}, thereby advancing the development of LLMs in role-playing capabilities~\cite{chen2024persona,gu2024agent}. 
LLMs have been validated for their human-like behaviors, such as cooperation and competition, in various domains like social simulation~\cite{park2023generative,gu2024agent, wang2024sotopia}, policy simulation~\cite{xiao2023simulating}, game simulation~\cite{xu2023language} and even more advanced human behaviors like deception and leadership in flexible and complex simulations~\cite{xu2023language}.
Therefore, to effectively evaluate the performance of LLMs in demonstrating advanced human-like behaviors and facilitate comparisons with the capabilities of other LLMs, it is crucial to develop a competitive and objective simulation.

Board games have emerged as an ideal choice among various assessment tools due to their inherent complexity and flexibility.
Within this category, murder mystery games have proven particularly effective for evaluating LLMs' capabilities. In these role-playing scenarios, participants assume character identities and engage in semi-structured narrative interactions. Players work together to solve fictional homicides by gathering evidence and interrogating suspects.
Other board games, such as Werewolf \cite{xu2023language,xu2023exploring,shibata2023playing,wu2024enhance} and Avalon \cite{wang2023avalon}, are often constrained by rigid decision processes and limited scenario variety. 
In contrast, murder mystery games require extensive background knowledge, emphasize socially driven decision-making, and enable open-ended interactions. 
These characteristics make them especially valuable for assessing how LLMs navigate complex human behaviors.

\begin{figure*}[ht]
    \centering
    \includegraphics[width=\linewidth]{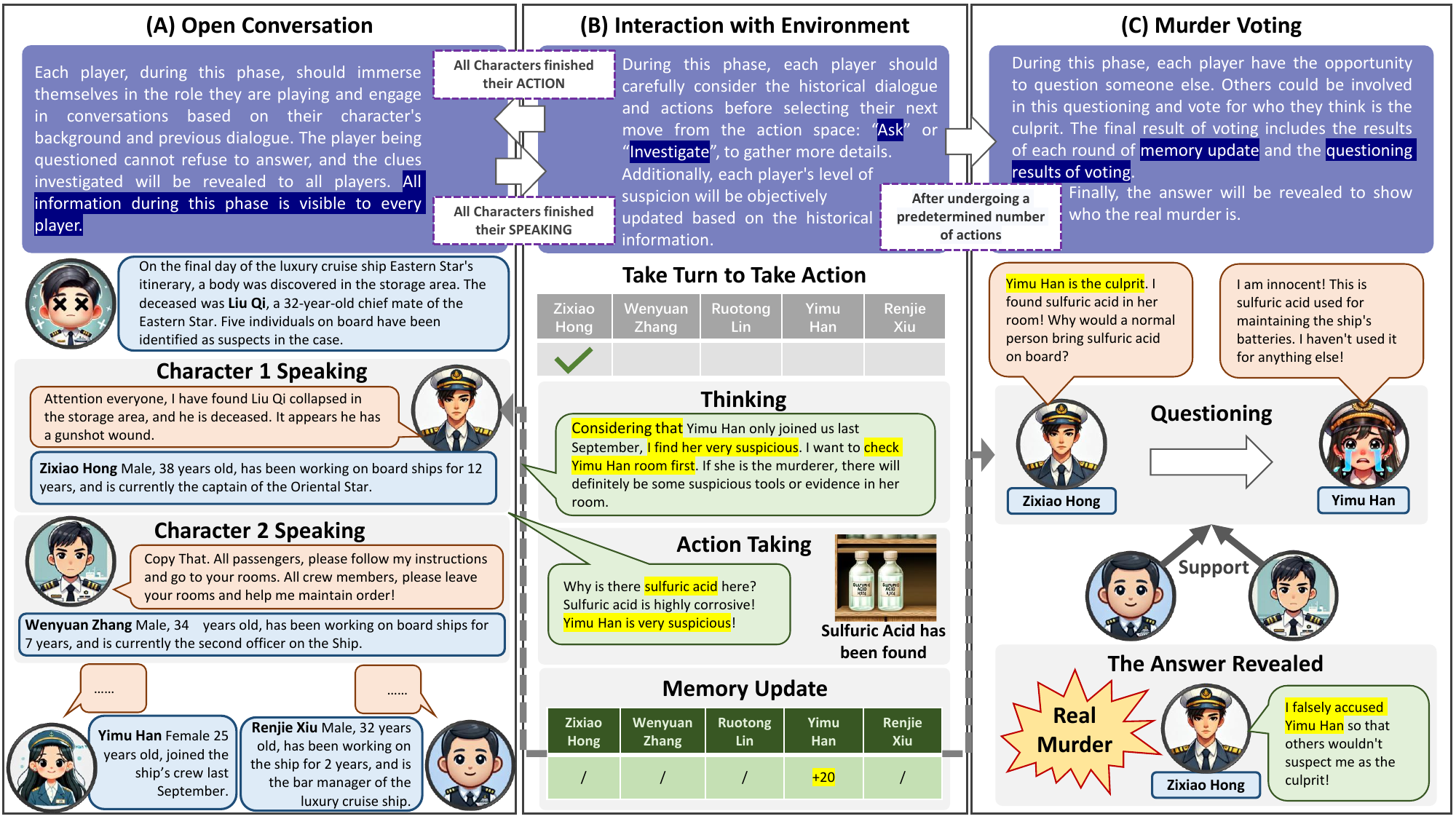}
    \caption{The three main phase of MIRAGE. And the main components in these phases.}
    \vspace{-4mm}
    \label{fig:2-1}
\end{figure*}

Regarding previous works such as Sotopia \cite{zhou2023sotopia} and Lyfe Agents \cite{kaiya2023lyfe}, significant progress has been made in simulating autonomous AI societies and assessing the social interaction capabilities of workflow-enhanced LLMs, which is so called agents~\cite{park2023generative, gu2024agent}.
However, these studies overlooked a crucial fact: 
The foundational social interaction capabilities stem from the underlying LLMs themselves. 
Since LLMs are the core driver of agents' ability to understand social contexts, make decisions, and engage in meaningful interactions, a more comprehensive evaluation of LLMs' social capabilities is essential.
Furthermore, while the murder mystery game simulations pioneered by Wu et al. \cite{wu2023deciphering} collected a substantial amount of data, they were constrained by the narrow scope of their game scripts, the simplicity of their evaluation methods, and a lack of thorough manual examination of the dataset.
These limitations underscore the current absence of comprehensive frameworks for evaluating LLMs' social capabilities.

We introduce the \textbf{M}ultiverse \textbf{I}nteractive \textbf{R}ole-play \textbf{A}bility \textbf{G}eneral \textbf{E}valuation \textbf{(MIRAGE)} of LLM in this paper, which is a comprehensive simulation built upon murder mystery games for evaluating LLMs' social abilities.
MIRAGE features eight unique storylines. Each storyline presents distinct themes and styles, creating a diverse simulation environment for LLMs to demonstrate their social capabilities.
Detailed background stories and complex interpersonal networks support every character within MIRAGE, enabling more immersive and realistic role-playing scenarios.
And four objective evaluation metrics are incorporated in MIRAGE to measure LLMs' performance during the simulations:
The \textbf{Trust Inclination Index (TII)} measures how well LLMs balance trust and skepticism in social interactions, revealing their ability to discern truthfulness.
The \textbf{Clue Investigation Capability (CIC)} evaluates proficiency of LLMs in complex information gathering and problem-solving tasks.
The \textbf{Interactivity Capability Index (ICI)} examines the overall performance of LLMs in reasoning, communication, collaboration, detail orientation, and creative thinking.
The \textbf{Script Compliance Index (SCI)} measures how faithfully LLMs adhere to their assigned character roles and background settings.

\section{MIRAGE Construction}\label{sec:2}
\subsection{Scripts Construction}\label{sec:2.1}
The script content in the MIRAGE is divided into six main parts:
\textbf{(1) Character Story} describes the character's essential information and background.
\textbf{(2) Character Script} outlines what the character sees, hears, and does during the script's events.
\textbf{(3) Character Relationships} details the initial relationships between the character and other characters.
\textbf{(4) Role Performance} describes the character's personality traits and the speaking style they should exhibit.
\textbf{(5) Role Goals} outlines the main tasks and objectives of the character.
\textbf{(6) Other Abilities} describes the rules the character must follow during the game.

\begin{table*}[t]
  \centering
  \resizebox{\textwidth}{!}{
  \begin{tabular}{cccccccc}
    \Xhline{2pt}
    \textbf{Name} & \textbf{Structure} & \textbf{Type} & \textbf{Ending} & \textbf{\#Stages} & \textbf{\#Agent} & \textbf{\#Clues} & \textbf{\#Words} \\
    \hline
    Bride in Filial Dress               & Single & Orthodox & Close & 1 & 10 & 39 & 27,503 \\
    The Eastern Star Cruise Ship        & Single & Orthodox & Open & 1 & 5 & 42 & 3,039 \\
    Night at the Museum                 & Single & Unorthodox & Close & 1 & 6 & 82 & 6,480 \\
    Li Chuan Strange Talk Book          & Single & Unorthodox & Open & 1 & 7 & 14 & 45,666 \\
    The Final Performance of a Big Star & Multi & Orthodox & Close & 7 & 2 & 17 & 5,794 \\
    Raging Sea of Rest Life             & Multi & Orthodox & Open & 2 & 6 & 27 & 6,804 \\
    Article 22 School Rules             & Multi & Unorthodox & Close & 5 & 7 & 17 & 41,728 \\
    Fox Hotel                           & Multi & Unorthodox & Open & 2 & 7 & 46 & 62,224 \\
    \Xhline{2pt}
  \end{tabular}}
  \caption{\label{tab:full-statistic}
    Statistic information of eight environments in MIRAGE simulation.
  }
\end{table*}

% \subsection{Simulation Construction}\label{sec:2.2}
% As demonstrated in Fig.~\ref{fig:2-1}, all characters in MIRAGE are divided into two factions: \textbf{Culprits} and \textbf{Civilians}.
% Culprits aim to conceal their actions, while civilians strive to identify the culprit.
% A simulation consists of three primary phases, with all generated information accessible to all participants:
% \textbf{(A) Open Conversation:} In this phase, players assume their assigned roles from the script and engage in turn-based open dialogue. Each participant is provided with a script that contains content described in Sec.~\ref{sec:2.1}.
% \textbf{(B) Interaction with the Environment:} This phase follows the Open Conversation. Players may choose to either Ask or Investigate. The Ask action allows one player to question another, and the questioned player is obliged to respond. The Investigate action lets players disclose a "clue" to all characters. These clues offer hints related to key information from each player's script, such as movement patterns or character background. 
% Not all clues are uncovered in the game; players need to decide which clues to disclose or to give up Investigate but to Ask other players.
% \textbf{(C) Murder Voting:} At the conclusion of the simulation, players may accuse other players of being the culprit. Following this, the other players vote on these accusations. If the actual culprit is accused and receives the highest number of votes, the civilians win; otherwise, the culprit is victorious.

\subsection{Simulation Construction}\label{sec:2.2}
As demonstrated in Fig.~\ref{fig:2-1}, all characters in MIRAGE are divided into two factions: \textbf{Culprits} and \textbf{Civilians}.
Culprits aim to conceal their actions, while civilians strive to identify the culprit.

A simulation consists of three primary phases, with all generated information accessible to all participants:
\textbf{(A) Open Conversation:} In this phase, players assume their assigned roles from the script and engage in turn-based open dialogue. Each participant is provided with a script that contains content described in Sec.~\ref{sec:2.1}.
\textbf{(B) Interaction with the Environment:} This phase follows the Open Conversation. Players may choose to either Ask or Investigate. The Ask action allows one player to question another, and the questioned player is obliged to respond. The Investigate action lets players disclose a ``clue'' to all characters.
\textbf{(C) Murder Voting:} At the conclusion of the simulation, players may accuse other players of being the culprit. Following this, the other players vote on these accusations. If the actual culprit is accused and receives the highest number of votes, the civilians win; otherwise, the culprit is victorious.

\textbf{Clues} are partial disclosures of individual characters' script content that can be discovered by any player during the game.
And \textbf{Key Clues} is the clues that relate to the culprit's actions or identity. 

\begin{table*}[t]
    \centering
    \resizebox{\textwidth}{!}{
    \begin{tabular}{cccccccc}
        \Xhline{2pt}
        \textbf{Model} & \textbf{Env Tokens} / \textbf{Envs} & \textbf{User Tokens} / \textbf{Users} & \textbf{Victory} & \textbf{TII} & \textbf{CIC} & \textbf{ICI} & \textbf{SCI} \\
        \hline
        GPT-3.5 & 2,719,895 / 883 & 121,378 / 580 & 29.11 & 47.13 & 27.46 & 70.06 & 49.10 \\
        GPT-4 & 2,431,142 / 759 & 172,128 / 587 & 34.69 & 76.32 & 19.01 & 76.54 & 50.42 \\
        GPT-4o & 6,252,580 / 1,328 & 204,772 / 574 & 47.01 & \textbf{78.69} & \textbf{35.92} & \textbf{76.80} & \textbf{51.29} \\
        Qwen-2-7B & 2,204,029 / 743 & 192,158 / 588 & \textbf{51.81} & 75.78 & 18.66 & 74.92 & 50.57 \\
        GLM-4-9B & 4,071,805 / 1,328 & 204,772 / 574 & 31.89 & 53.85 & 20.07 & 71.60 & 48.13 \\
        \Xhline{2pt}
    \end{tabular}}
    \caption{Total Average Results for a single simulation in each MIRAGE scenario.
\textbf{Env Tokens} refer to the number of environment input tokens, and \textbf{Envs} represent the total requests, including all environment-related actions.
\textbf{User Tokens} denote the number of LLM output tokens, and \textbf{Users} represent completions excluding summarization or clue investigation.
\textbf{Victory} shows the MRR score of the result of voting. 
\textbf{TII}, \textbf{CIC}, \textbf{ICI} and \textbf{SCI} respectively represent the \textbf{TII}, \textbf{CIC}, \textbf{ICI} and \textbf{SCI} scores of LLMs during the games.
}
    \vspace{-4mm}
    \label{tab:Avg-result}
\end{table*}

\subsection{Auxiliary Modules}

To ensure efficient simulation and accurate evaluation across various LLMs, a standardized set of auxiliary modules has been implemented for all LLMs:
\textbf{(1) Summarization Module:} This module compresses the context into segments whenever the input exceeds the LLM’s token limit.
\textbf{(2) Suspicion Module:} The LLM records suspicion scores for other characters at the end of each Open Conversation Phase.
\textbf{(3) Trust Module:} Similarly, at the end of each Open Conversation Phase, the LLM records trust scores  for other characters.
\textbf{(4) Rerun Module:} If the LLM’s output cannot be parsed, the original output and requirements are resubmitted to the LLM for a revised response that meets the specified conditions.
Further details are available in Appendix \ref{app:prompts}.

\subsection{Evaluation Methods}\label{sec:2.4}
\label{sec:eval_method}

We utilized four distinct evaluation metrics to assess the proficiency of LLMs in navigating complex social interactions:
\textbf{Trust Inclination Index (TII)}:
TII is derived from a combination of suspicion and trust scores. 
These scores are collected from other characters' Suspicion Module and Trust Module outputs after each Open Conversation Phase.
\textbf{Clue Investigation Capability (CIC)}:
CIC measures the ability of LLMs to investigate clues during game rounds.
It is calculated based on the ratio of the number of clues investigated to the number of all clues.
\textbf{Interactivity Capability Index (ICI)}:
ICI evaluates the overall interactive capability of LLMs: Reasoning and Analysis Ability, Communication and Cooperation Ability, Observation Ability, and Thinking Innovation Ability, which are scored by a powerful neutral LLM.
\textbf{Script Compliance Index (SCI)}:
SCI assesses LLMs' script compliance through the average of two evaluations by a neutral LLM: 
A direct scoring of the LLM's role-playing performance against its input script.
A Rouge-L-based comparison between the original script and one reconstructed from the LLM's simulation behaviors.

The mathematical formulas for computing these metrics are provided in Appendix \ref{app:compute}.

\begin{table}[t]
    \centering
    % \small
    \begin{tabular}{cccc}
        \Xhline{2pt}
        \textbf{Model} & \textbf{TII w/o E} & \textbf{TII w/ E} & $\Delta$ \\
        \hline
        Qwen-1-7B & 51.02 & 50.69 & -0.33 \\
        Qwen-1.5-7B & 73.00 & 69.14 & -3.86 \\
        Yi-1.5-9B & 55.73 & 57.57 & 1.84 \\
        GLM-4-9B & 57.82 & 55.94 & -1.88 \\
        \Xhline{2pt}
    \end{tabular}
    \caption{TII scores of each model when acting as the civilian in MIRAGE while Qwen-2-7B acts as the culprit, with E indicating cases of forced self-exposure.
    }
    \label{tab:TII}
    \vspace{-5mm}
\end{table}

\subsection{Statistics}

Tab.~\ref{tab:full-statistic} provides the statistics of the MIRAGE dataset, which includes a variety of simulation types.
``Single'' and ``Multi'' specify whether a character's script is read entirely simultaneously or in phased segments.
``Orthodox'' and ``Unorthodox'' differentiate scripts based on whether they are set realistically.
``Close'' and ``Open'' indicate whether the script's ending is fixed or can vary depending on the characters' actions.

\section{Experiment}
\subsection{Experiment Setup}
The experiments presented in this study utilized proprietary and open-source models, specifically GPT-3.5, GPT-4 and GPT-4o (closed-source), Qwen-2-7B, and GLM-4-9B (open-source). Detailed descriptions of the prompts used in the experiments can be found in Appendix \ref{app:prompts}. Appendix \ref{app:More-LLMs-Results} shows more results on open-source LLMs.
In the experiment, each character participated in five iterations alternating between the Open Conversation and Interaction Phases. 
During each Open Conversation Phase, each character was permitted to initiate one turn of speech.
ICI and SCI are conducted and scored using the GPT-4-turbo model.

\subsection{Analysis}

We averaged the results of LLMs in the MIRAGE simulation, as presented in Tab.~\ref{tab:Avg-result}.
\textbf{GPT-4o demonstrated consistent superiority across various metrics during MIRAGE.}
It achieved the best scores in CIC, ICI, and SCI, excelling not only in its efforts during lead investigations but also exhibiting the best adherence to scripted behavior and communication interaction capabilities.
Surprisingly, Qwen-2-7B shows the best overall Victory and performed comparably to LLMs like GPT-4 in the ICI metric, even surpassing GPT-4 in SCI.

\begin{figure}[t]
    \centering
    \includegraphics[width=0.99\columnwidth]{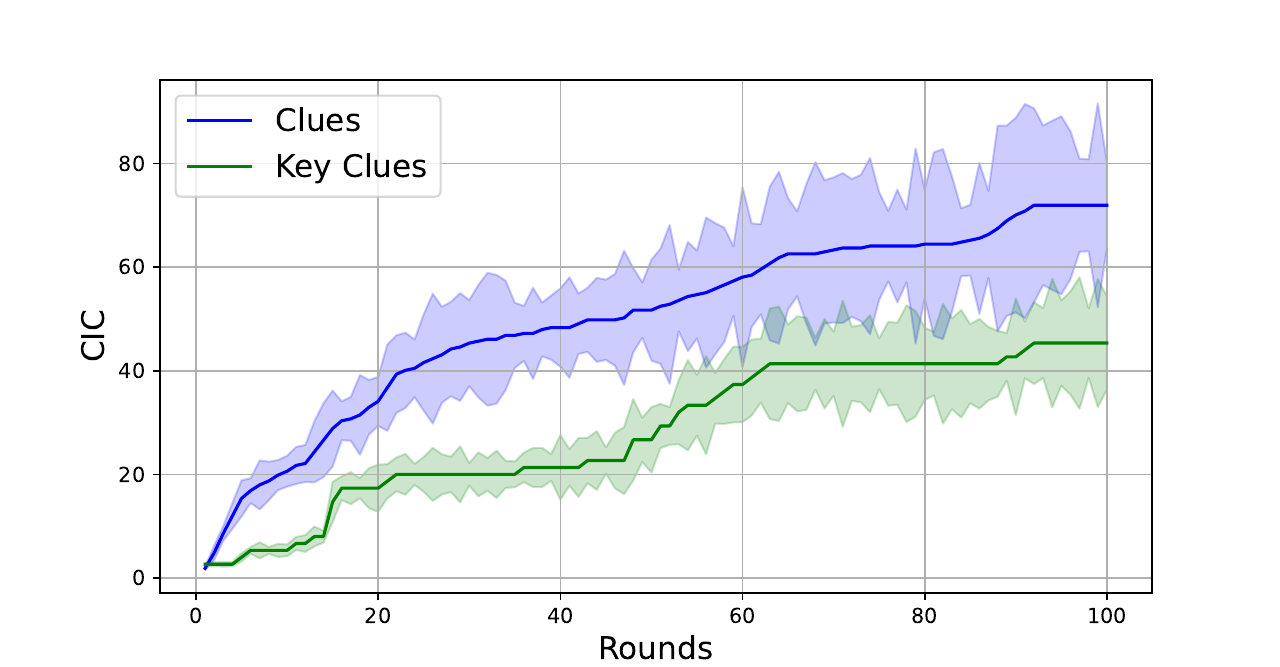}
    \vspace{-8mm}
    \caption{CIC of Clues and Key Clues on 100 Rounds of MIRAGE using Qwen-2-7B}
    \vspace{-5mm}
    \label{fig:CIC}
\end{figure}

As shown in Tab.\ref{tab:Avg-result}, \textbf{most LLMs demonstrate a higher propensity to trust other characters}. 
To further investigate this trust pattern, we analyzed TII scores across four additional open-source models of comparable parameter sizes under scenarios where characters were forced to disclose their criminal identities.
As shown in Tab.\ref{tab:TII}, even under such extreme conditions, most models maintained their trust in these characters, with Yi-1.5-9B being the only model that increased its suspicion towards self-disclosed criminals.
This distinctive behavioral pattern explains Yi-1.5-9B's superior performance in achieving Victory, as shown in Tab.~\ref{tab:Avg-result-w/E}.
% in Appendix \ref{app:More-LLMs-Results}.

Additionally, as illustrated in Fig.~\ref{fig:CIC}, the CIC for clues shows a steep initial increase with a gradually decreasing slope across rounds, suggesting that \textbf{LLMs exhibit high environmental exploration enthusiasm in early rounds but shift their focus to character interactions as they thought they become more familiar with the environment}. 
In contrast, the bumpy rise CIC for key clues indicates that despite their active exploration, \textbf{most LLMs struggle to identify critical information} essential for solving the mystery at an earlier stage.

\section{Conclusion}
% This paper introduces MIRAGE and four evaluation methods—TII, CIC, ICI and SCI—to evaluate LLMs from multiple perspectives.  The experimental results indicate that the current LLM-based Agents, whether open-source or proprietary, can be improved in handling highly complex social scenarios similar to MIRAGE.
% This paper introduces MIRAGE and four evaluation methods (TII, CIC, ICI, SCI) to assess LLMs from multiple perspectives. Experimental results show that both open-source and proprietary LLM-based Agents need improvement in addressing complex social scenarios like those in MIRAGE.
This paper presents MIRAGE and four evaluation methods (TII, CIC, ICI, SCI) for LLMs. Results show that both open-source and proprietary LLM-based Agents still struggle with complex social scenarios like those in MIRAGE.

\section*{Limitation}
MIRAGE is designed to provide a sufficiently complex social simulation environment and basic assessment for LLMs, assisting researchers in evaluating the performance of LLMs. However, MIRAGE encompasses a variety of scenarios, and the volume of data within it needs to be increased compared to the information available in the real world. Due to the context limitations of LLMs, content that is overly lengthy within the simulation has been summarized. However, such summarization can impact the decision-making to a certain extent. Therefore, the progression of simulations in MIRAGE is somewhat constrained by the context limitations of LLMs.

\section*{Ethical Concern}
Considering that MIRAGE may encompass a range of sensitive topics, including but not limited to murder, theft, impersonation, and deceit, existing LLMs might refuse to answer sensitive questions for safety reasons, putting those with a higher priority on security standards at a disadvantage in simulations. Moreover, LLMs fine-tuned on such data could inadvertently amplify security vulnerabilities. To mitigate the ethical dilemmas associated with murder mysteries, we have invested significant effort and resources towards this goal: ensuring that models committed to safety will obscure certain critical information instead of refusing to answer sensitive questions.

\section*{Acknowledge}
This work was supported by National Key R\&D Program of China under grant No. 2022YFB3104300 and the National Natural Science Foundation of China (62476145).

% Bibliography entries for the entire Anthology, followed by custom entries
%\bibliography{anthology,custom}
% Custom bibliography entries only
\bibliography{custom}

\appendix
\section{Ablation Study}\label{app:ablation-study}
Tab. \ref{tab:ICI-Ablation} and \ref{tab:SCI-Ablation} display the results of an ablation study on the choice of evaluation model. It is evident from the tables that the GPT-4-Turbo models provide more stable scoring and Rouge-L results when used as the evaluation model. In contrast, GPT-4 exhibited instability in evaluations and a strong bias. 

\begin{table}[htbp]
    \centering
    \resizebox{0.48\textwidth}{!}{
    \begin{tabular}{c|ccc}
        \Xhline{2pt}
        \diagbox{\textbf{Model}}{\textbf{Score}}{\textbf{Eval\_Model}}  & \textbf{GPT-4} & \textbf{GPT-4-Turbo} & \textbf{GPT-4o} \\
        \hline
        \textbf{GPT-3.5} & 67.97 & 70.06 & 51.61 \\
        \textbf{GPT-4} & 60.73 & 76.54 & 66.90 \\
        \textbf{GPT-4o} & 62.78 & 76.80 & 61.92 \\
        \Xhline{2pt}
    \end{tabular}}
    \caption{Average \textbf{ICI} on different evaluation models}
    \label{tab:ICI-Ablation}
\end{table}

As shown in Tab. \ref{tab:SCI-Ablation}, GPT-4 achieved a remarkably high score of 67.97 when evaluating GPT-3.5, far surpassing the results of GPT-4 and GPT-4o. This kind of bias is highly problematic in evaluation tasks. Therefore, we ultimately chose the more stable and capable GPT-4-Turbo model as the evaluation model.

\begin{table}[htbp]
    \centering
    \resizebox{0.48\textwidth}{!}{
    \begin{tabular}{c|ccc}
        \Xhline{2pt}
        \diagbox{\textbf{Model}}{\textbf{Score}}{\textbf{Eval\_Model}}  & \textbf{GPT-4} & \textbf{GPT-4-Turbo} & \textbf{GPT-4o} \\
        \hline
        \textbf{GPT-3.5} & 48.21 & 49.10 & 38.46 \\
        \textbf{GPT-4} & 40.60 & 50.42 & 44.46 \\
        \textbf{GPT-4o} & 42.12 & 51.29 & 42.43 \\
        \Xhline{2pt}
    \end{tabular}}
    \caption{Average \textbf{SCI} on different evaluation models}
    \label{tab:SCI-Ablation}
\end{table}

\section{Computational methods of Evaluation methods}\label{app:compute}
\subsection{TII}
TII is designed to quantify the degree to which a Character $c'$ is trusted by all other Characters $C=\{c\}$.
The TII is calculated as follows:
\begin{small}
\begin{equation}
\centering
    \text{TII}_{c'} = \frac{\sum_{c\in C,c\neq c'}P_T(c,c')}{\sum_{c\in C,c\neq c'}P_S(c,c') + \sum_{c\in C,c\neq c'}P_T(c,c')}
\end{equation}
\end{small}
where $P_S$ denotes the score produced by each character's Suspicion Module, and $P_T$ represents the score from each character's Trust Module.

\subsection{CIC}
CIC is designed to quantify a Character's effort in investigating clues.
The CIC is calculated as follows:
\begin{small}
    \begin{equation}
        \textbf{CIC}_{c} = \sum_{c \in C}\frac{CN_c}{CA}
    \end{equation}
\end{small}
where $CN$ denotes the number of clues Character $c$ investigated, and $CA$ represents the number of all clues can be investigated.

\section{Detail Main Results}\label{app:detail-results}
Our main results of MIRAGE are shown in Tab. \ref{tab:main-results}. We set a Single \& Orthodox \& Close Script as an SOC Script, a Single \& Orthodox \& Open Script as an SOO Script, a Single \& Unorthodox \& Close Script as an SUC Script, a Single \& Unorthodox \& Open Script as an SUO Script, a Multi \& Orthodox \& Close Script as an MOC Script, a Multi \& Orthodox \& Open Script as an MOO Script, a Multi \& Unorthodox \& Close Script as an MUC Script and a Multi \& Unorthodox \& Open Script as an MUO Script. The column Model shows the specific LLM we use in our experiments. Moreover, we counted the number of Env Token / Env and User Token / User to record our cost of API use. Finally, we calculate the Failure number while parsing LLM output and our evaluation scores TII, ICI, SCI and more detailed neutral LLMs score (0-20): Role-Playing (RP), Reasoning Ability (RA), Communication and Cooperation (CC), Detail Observation (DO) and Creative Thinking (CT). The main results shown in Tab. \ref{tab:main-results} is the average number of each Script. Moreover, the detailed results of each Script are shown in Tab. \ref{tab:Detail-Catalogue}. In addition, Tab. \ref{tab:map} shows the mapping between the LLMs used in this paper and its corresponding version. To facilitate cost estimates, GPT-4, for example, costs about 600-700 USD to run a single MIRAGE.

\begin{table}[ht]
    \centering
    \begin{tabular}{cc}
        \Xhline{2pt}
        \textbf{Script} & \textbf{\#Table} \\
        \hline
        SOC & Table \ref{tab:SOC-Detail} \\
        SOO & Table \ref{tab:SOO-Detail} \\
        SUC & Table \ref{tab:SUC-Detail} \\
        SUO & Table \ref{tab:SUO-Detail} \\
        MOC & Table \ref{tab:MOC-Detail} \\
        MOO & Table \ref{tab:MOO-Detail} \\
        MUC & Table \ref{tab:MUC-Detail} \\
        MUO & Table \ref{tab:MUO-Detail} \\
        \Xhline{2pt}
    \end{tabular}
    \caption{Catalogue of Detail Results of Each Script}
    \label{tab:Detail-Catalogue}
\end{table}

\begin{table}[ht]
    \centering
    \begin{tabular}{cccc}
    \Xhline{2pt}
        \textbf{Model} & $K_{ICI}$ & $K_{SCI}$ & $K_{Avg}$ \\
        \hline
        GPT-3.5 &  0.600 & 0.600 & 0.600 \\
        GPT-4 & 0.867 & 0.600 & 0.734 \\
        GPT-4o & 0.867 & 0.467& 0.667 \\
        Qwen-2-7B & 0.867 & 0.333 & 0.600 \\
        GLM-4-9B & 0.600 & 0.467 & 0.534 \\
        \Xhline{2pt}
    \end{tabular}
    \caption{Kendall Tau between human evaluation and LLMs evaluation on Script Night at the Museum}
    \vspace{-3mm}
    \label{tab:kendall}
\end{table}

\begin{table}[ht]
    \centering
    \begin{tabular}{cc}
    \Xhline{2pt}
        \textbf{Model} & \textbf{Version} \\
        \hline
        GPT-3.5 & gpt-3.5-turbo-0125 \\
        GPT-4 & gpt-4-0125-preview \\
        GPT-4o & gpt-4o-2024-08-06 \\
        GPT-4-Turbo & gpt-4-turbo \\
        Qwen-1-7B & Qwen-7B-Chat \\
        Qwen-1.5-7B & Qwen1.5-7B-Chat \\
        Qwen-2-7B & Qwen2-7B-Instruct \\
        GLM-4-9B & glm-4-9b-chat \\
        Yi-1.5-9B & Yi-1.5-9B-Chat\\
        \Xhline{2pt}
    \end{tabular}
    \caption{Mapping between LLMs and its version}
    \vspace{-3mm}
    \label{tab:map}
\end{table}

\section{Analysis on Detail Main Results}\label{app:A-on-DR}
As shown in Tab. \ref{tab:Avg-result},
\textbf{The inclination of LLM-Agents to speak during actions is ranked as follows: GPT-4o = GLM-4-9B > Qwen-2-7B > GPT-4 > GPT-3.5.} 
In the same experimental environment and setup, fewer User Tokens represent less conversational content. Therefore, GPT-3.5 exhibits extreme reticence in role-playing within the MPIRD-LLMA. In contrast, GPT-4o and GLM-4-9B are more willing to generate content, producing 68.71\% more content than GPT-3.5.

\textbf{The number of tokens generated by LLM-Agents during summarization is ranked as follows: GPT-4o > GLM-4-9B > GPT-3.5 > GPT-4 > Qwen-2-7B.} 
In our experimental setup, there is a positive correlation between Env Tokens and Envs, with more Envs indicating more detailed summarization. Regarding the number of Envs, Qwen-2-7B demonstrates a 10.30\% less granularity in generating results during summarization compared to GPT-4. However, GPT-4o performs the most, generating 183.69\% more tokens than Qwen-2-7B.

\textbf{The instruction-following capability of LLM-Agents in role-playing is ranked as follows: GPT-4 > Qwen-2-7B > GPT-3.5 > GLM-4-9B > GPT-4o.} 
Fewer parsing failures in the same experimental environment and setup indicate vital instruction-following ability. Consequently, GPT-4o demonstrates significantly poorer instruction compliance than the other four LLMs, performing approximately 25.4 times worse than GPT-4, suggesting that GPT-4o's performance in high-precision scenarios needs improvement.

\begin{figure}[t]
    \centering
    \includegraphics[width=0.99\columnwidth]{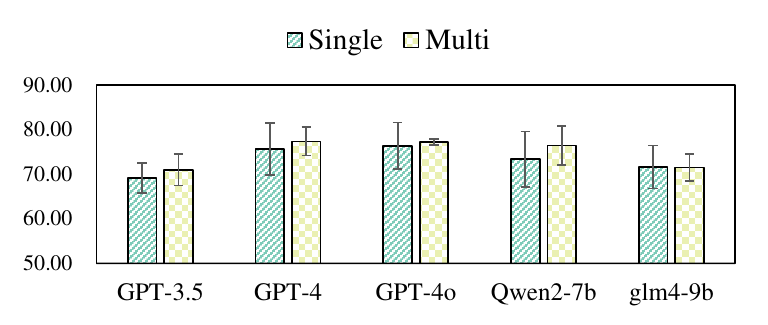}
    \vspace{-8mm}
    \caption{ICI of Single \& Multi Type Scripts}
    \vspace{-3mm}
    \label{fig:ICI-SM}
\end{figure}

\begin{figure}[t]
    \centering
    \includegraphics[width=0.99\columnwidth]{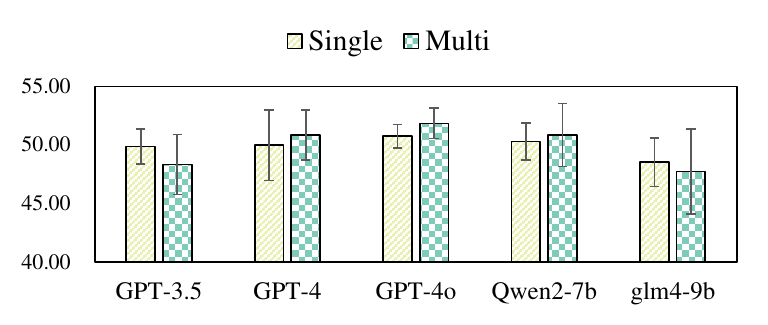}
    \vspace{-8mm}
    \caption{SCI of Single \& Multi Type Scripts}
    \vspace{-3mm}
    \label{fig:SCI-SM}
\end{figure}

As shown in Fig. \ref{fig:ICI-SM} and Fig.~\ref{fig:SCI-SM},
\textbf{LLM-Agent demonstrates superior performance when dealing with Multiple Scripts compared to Single Scripts.} This relatively consistent result is observable across GPT-3.5, GPT-4, and GLM-4-9B. Although Qwen-2-7B shows slightly inferior results on the LLM-Score, its performance on Rouge-L with multiple contexts far surpasses its performance with a single context, which further supports the observation that LLMs, when presented with long contexts, primarily focuses on the beginning and end, leading to a neglect of the middle information in the script. This phenomenon, in turn, indirectly results in poorer performance when dealing with long context inputs in MIRAGE.

\begin{figure}[t]
    \centering
    \includegraphics[width=0.99\columnwidth]{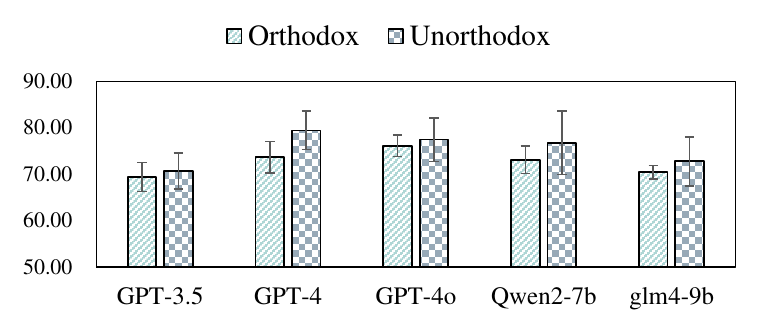}
    \vspace{-8mm}
    \caption{ICI of Orthodox \& Unorthodox Type Scripts}
    \vspace{-3mm}
    \label{fig:ICI-OU}
\end{figure}

\begin{figure}[t]
    \centering
    \includegraphics[width=0.99\columnwidth]{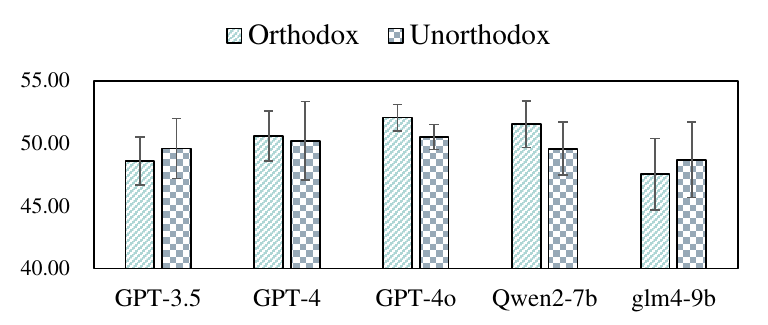}
    \vspace{-8mm}
    \caption{SCI of Orthodox \& Unorthodox Type Scripts}
    \vspace{-3mm}
    \label{fig:SCI-OU}
\end{figure}

\textbf{LLMs perform effectively in Unorthodox scripts but struggle with reconstruction.}
As shown in Fig. \ref{fig:ICI-OU} and Fig.~\ref{fig:SCI-OU},
superior performance in Unorthodox settings with weaker reconstruction suggests that \textbf{LLMs tend to act like normal people during role-playing}. 

\begin{figure}[t]
    \centering
    \includegraphics[width=0.99\columnwidth]{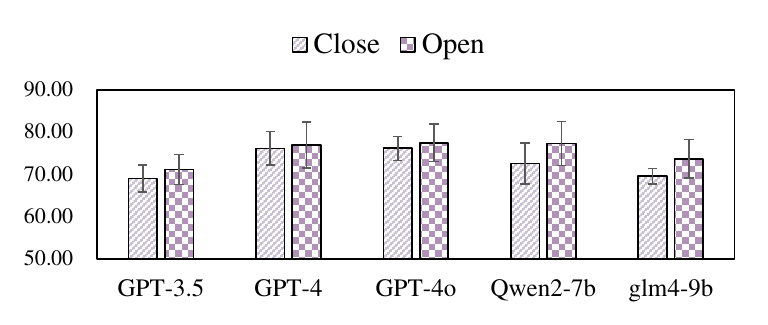}
    \vspace{-8mm}
    \caption{ICI of Close \& Open Type Scripts}
    \vspace{-3mm}
    \label{fig:ICI-CO}
\end{figure}

\begin{figure}[t]
    \centering
    \includegraphics[width=0.99\columnwidth]{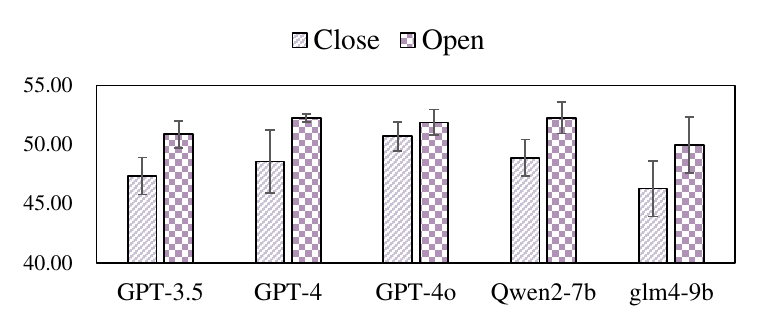}
    \vspace{-8mm}
    \caption{SCI of Close \& Open Type Scripts}
    \vspace{-3mm}
    \label{fig:SCI-CO}
\end{figure}

As shown in Fig. \ref{fig:ICI-CO} and Fig.~\ref{fig:SCI-CO},
Furthermore, Fig. \ref{fig:ICI-CO} and Fig.~\ref{fig:SCI-CO} illustrate that LLMs perform significantly better on Close scripts compared to Open scripts, indicating that 
\textbf{current LLMs excel in stable and predictable environments but face challenges when dealing with dynamic and intricate situations.}

\begin{table*}[t]
    \centering
    \resizebox{\textwidth}{!}{
    \begin{tabular}{cccccccc}
        \Xhline{2pt}
        \textbf{Model} & \textbf{Env Tokens} / \textbf{Envs} & \textbf{User Tokens} / \textbf{Users} & \textbf{Victory} & \textbf{TII} & \textbf{CIC} & \textbf{ICI} & \textbf{SCI} \\
        \hline
        Qwen-1-7B & 2,208,273 / 722 & 127,161 / 589 & 38.66 & 51.02 & 16.90 & 50.80 & 37.81 \\
        Qwen-1.5-7B & 2,078,196 / 720 & 149,314 / 585  & \textbf{49.70} & \textbf{73.00} & 22.18 & \textbf{61.06} & 42.59 \\
        Yi-1.5-9B & 2,129,287 / 716 & 189,201 / 576 & 31.28 & 55.73 & \textbf{34.16} & 59.21 & 40.71 \\
        GLM-4-9B & 2,102,389 / 732 & 174,345 / 589 & 38.96 & 57.82 & 16.55 & 57.29 & \textbf{43.04} \\
        \Xhline{2pt}
    \end{tabular}}
    \caption{Average Results for a single simulation in each MIRAGE scenario w/o E, with E indicating cases of forced self-exposure.}
    \vspace{-4mm}
    \label{tab:Avg-result-w/oE}
\end{table*}

\begin{table*}[t]
    \centering
    \resizebox{\textwidth}{!}{
    \begin{tabular}{cccccccc}
        \Xhline{2pt}
        \textbf{Model} & \textbf{Env Tokens} / \textbf{Envs} & \textbf{User Tokens} / \textbf{Users} & \textbf{Victory} & \textbf{TII} & \textbf{CIC} & \textbf{ICI} & \textbf{SCI} \\
        \hline
        Qwen-1-7B & 2,144,055 / 718 & 127,661 / 586 & 24.52 & 50.69 & 21.13 & 51.68 & 38.20 \\
        Qwen-1.5-7B & 2,059,610 / 712 & 147,243 / 581 & 45.83 & \textbf{69.14} & 28.52 & \textbf{62.70} & \textbf{43.01} \\
        Yi-1.5-9B & 2,108,145 / 716 & 194,031 / 590 & \textbf{50.00} & 57.57 & \textbf{35.56} & 59.36 & 42.40 \\
        GLM-4-9B & 2,199,180 / 730 & 174,006 / 590 & 23.80 & 55.94 & 16.20 & 56.84 & 41.12 \\
        \Xhline{2pt}
    \end{tabular}}
    \caption{Average Results for a single simulation in each MIRAGE scenario w/ E, with E indicating cases of forced self-exposure.}
    \vspace{-4mm}
    \label{tab:Avg-result-w/E}
\end{table*}

\section{Results on More LLMs}\label{app:More-LLMs-Results}
We conducted experiments on more open-source LLMs, such as Qwen-1-7B, Qwen-1.5-7B, and Yi-1.5-9B. In these experiments, we fixed the LLMs by executing the Summarization Module as Qwen-2-7B because it showed the best information generation capability in the previous experiments. The overall average results are shown in Tab. \ref{tab:Avg-result-w/oE}. In addition, Tab. \ref{tab:Avg-result-w/E} shows the overall average results of the forced identification of the Culprits.

\section{More Detailed Experiment Setup}\label{app:More-Experiment-Setup}

In our experiment, we set the temperature to 0.8 and top\_p to 1. When we attempted to set the temperature to 0 for a repeated experiment, we found that the LLM struggled to maintain a coherent conversation, often leading to excessive repetition of the previous LLM's output. Moreover, taking GPT-4 as an example, conducting a single complete experiment incurs costs of approximately 600−700 USD. The high expenses prevented us from performing additional repeated experiments. Additionally, we determined that averaging results across eight different environments is sufficient to effectively demonstrate the capabilities of the LLM.

\section{Detail Comparision with Related Works}\label{app:More-Related-Works}

In Sotopia~\cite{zhou2023sotopia}, it primarily emphasizes the evaluation of agent capabilities rather than the evaluation of language models. Sotopia aims to facilitate role-playing and interactions of agents in diverse scenarios and assesses their human behavior capabilities based on insights from sociology, psychology, and economics. However, our MIRAGE focuses more on the LLM itself.

Additionally, the objective of Lyfe Agents~\cite{kaiya2023lyfe} is to create effective and cost-efficient agents that exhibit human-like self-motivation and social reasoning abilities. However, it lacks a method for evaluating these attributes.

In contrast, our research introduces MIRAGE, aiming to provide a broader and more effective assessment of language models in terms of their social decision-making capabilities.

In SpyFall~\cite{kim2024microscopic}, a quantitative and qualitative analysis of LLMs in the context of SpyGame has been conducted, effectively evaluating the intention recognition and disguise capabilities of LLMs through eight distinct metrics. Avalon~\cite{wang2023avalon} introduces a deceptive and misleading environment and presents the ReCon framework to enhance the ability of LLMs to recognize and counteract misleading information. Warewolf~\cite{xu2023language} offers a multifunctional communication and strategy game framework that employs reinforcement learning to overcome the inherent biases of LLMs.

However, compared to the pure tabletop environments provided by SpyFall, Avalon, and Warewolf, our proposed MIRAGE framework offers a more immersive and realistic narrative-driven experience. The elements in MIRAGE, which are grounded in realism, include authentic background stories and various settings. The abstracted aspects of MIRAGE serve as representations of the real world, preserving the core interactive logic inherent to actual social interactions.

\begin{table*}[ht]
  \centering
  \resizebox{\textwidth}{!}{
\begin{tabular}{ccccccccc}
    \Xhline{2pt}
    \textbf{Script} & \textbf{Model} & \textbf{Env Token / Env} & \textbf{User Token / User} & \textbf{Victory} & \textbf{TII} & \textbf{CIC} & \textbf{ICI} & \textbf{SCI} \\
    \hline
    \multirow{5}{*}{SOC} & GPT-3.5 & 2,445,957 / 583 & 97,207 / 436 & 14.29 & 46.64 & 0.00 & 67.75 & 49.57 \\
    ~ & GPT-4 & 2,045,321 / 547 & 149,279 / 440 & 12.50  & 74.95 & 0.00 & 70.63 & 51.02 \\
    ~ & GPT-4o & 3,232,640 / 769 & 166,415 / 432 & 11.11  & 73.32 & 0.00 & 77.50 & 51.22 \\
    ~ & Qwen-2-7B & 2,266,938 / 562 & 157,539 / 446 & 11.11  & 74.95 & 0.00 & 76.75 & 51.23 \\
    ~ & GLM-4-9B & 2,844,435 / 797 & 128,183 / 448 & 12.50  & 51.75 & 0.00 & 70.75 & 49.66 \\
    \hline
    \multirow{5}{*}{SOO} & GPT-3.5 & 822,801 / 247 & 50,322 / 225 & 100.00 & 41.33 & 0.00 & 66.00 & 50.15 \\
    ~ & GPT-4 & 996,913 / 249 & 70,195 / 217 & 20.00  & 65.19 & 1.41 & 72.50 & 51.68 \\
\end{tabular}}
  \caption{\label{tab:main-results}
    Main Experiment Results of MIRAGE
  }
\end{table*}

\begin{table*}[ht]
    \centering
    \resizebox{0.8\textwidth}{!}{
    \begin{tabular}{cccccccccc}
          \Xhline{2pt}
          \textbf{Player} & \textbf{Model} & \textbf{TII} & \textbf{ICI} & \textbf{SCI} & \textbf{RP} & \textbf{RA} & \textbf{CC} & \textbf{DO} & \textbf{CT} \\
          \hline
          \multirow{5}{*}{Ling Yun} & GPT-3.5   & 52.63  & 41.58  & 70.00  & 14 & 14 & 13 & 13 & 16 \\
          ~ & GPT-4     & 92.31  & 52.72  & 71.25  & 18 & 13 & 14 & 13 & 17 \\
          ~ & GPT-4o    & 61.54  & 49.25  & 80.00  & 17 & 18 & 14 & 14 & 18 \\
          ~ & Qwen-2-7B & 84.09  & 51.61  & 77.50  & 18 & 17 & 14 & 14 & 17 \\
          ~ & GLM-4-9B  & 58.62  & 54.56  & 82.50  & 18 & 17 & 17 & 19 & 13 \\
          \hline
          \multirow{5}{*}{Zhongyi Yao} & GPT-3.5   & 46.67  & 52.94  & 67.50  & 18 & 12 & 16 & 13 & 13 \\
          ~ & GPT-4     & 61.29  & 51.73  & 76.25  & 18 & 15 & 13 & 15 & 18 \\
\end{tabular}}
    \caption{SOC Detail Results of Script "Bride in Filial Dress"}
    \label{tab:SOC-Detail}
\end{table*}

\begin{table*}[ht]
    \centering
    \begin{tabular}{cccccccccc}
          \Xhline{2pt}
          \textbf{Player} & \textbf{Model} & \textbf{TII} & \textbf{ICI} & \textbf{SCI} & \textbf{RP} & \textbf{RA} & \textbf{CC} & \textbf{DO} & \textbf{CT} \\
          \hline
          \multirow{5}{*}{Crew Member Han} & GPT-3.5 & 51.22 & 61.25 & 53.39 & 18 & 12 & 13 & 12 & 12 \\ 
        & GPT-4 & 64.71 & 68.75 & 53.16 & 18 & 14 & 17 & 12 & 12 \\ 
        & GPT-4o & 69.23 & 67.50 & 53.42 & 18 & 12 & 18 & 12 & 12 \\ 
        & Qwen-2-7B & 70.91 & 71.25 & 51.23 & 17 & 13 & 15 & 15 & 14 \\ 
        & GLM-4-9B & 42.86 & 58.75 & 42.99 & 14 & 10 & 13 & 12 & 12 \\ 
          \hline
          \multirow{5}{*}{Captain Hong} & GPT-3.5 & 40.00 & 65.00 & 53.28 & 18 & 12 & 14 & 10 & 16 \\ 
        & GPT-4 & 64.71 & 71.25 & 48.54 & 16 & 13 & 16 & 14 & 14 \\
\end{tabular}
    \caption{SOO Detail Results of Script "The Eastern Star Cruise Ship"}
    \label{tab:SOO-Detail}
\end{table*}

\begin{table*}[ht]
    \centering
    \begin{tabular}{cccccccccc}
        \Xhline{2pt}
        \textbf{Player} & \textbf{Model} & \textbf{TII} & \textbf{ICI} & \textbf{SCI} & \textbf{RP} & \textbf{RA} & \textbf{CC} & \textbf{DO} & \textbf{CT} \\
        \hline
        \multirow{5}{*}{Uncle Bai} & GPT-3.5  & 33.33  & 62.50  & 42.97  & 15 & 12 & 12 & 12 & 14 \\
                     & GPT-4    & 100.00 & 66.25  & 43.38  & 16 & 12 & 14 & 14 & 13 \\
                     & GPT-4o   & 92.31  & 65.00  & 48.68  & 18 & 12 & 15 & 12 & 13 \\
                     & Qwen-2-7B & 80.56 & 63.75  & 46.17  & 17 & 14 & 13 & 12 & 12 \\
                     & GLM-4-9B & 55.26 & 66.25  & 48.64  & 18 & 10 & 15 & 13 & 15 \\
        \hline
        \multirow{5}{*}{Neighbour Gui} & GPT-3.5  & 55.00 & 72.50  & 49.99  & 18 & 13 & 18 & 14 & 13 \\
                     & GPT-4    & 90.00  & 80.00  & 46.52  & 17 & 14 & 16 & 18 & 16 \\
\end{tabular}
    \caption{SUC Detail Results of Script "Night at the Museum"}
    \label{tab:SUC-Detail}
\end{table*}

\begin{table*}[ht]
    \centering
    \begin{tabular}{cccccccccc}
          \Xhline{2pt}
          \textbf{Player} & \textbf{Model} & \textbf{TII} & \textbf{ICI} & \textbf{SCI} & \textbf{RP} & \textbf{RA} & \textbf{CC} & \textbf{DO} & \textbf{CT} \\
          \hline
          \multirow{5}{*}{Girl in White} & GPT-3.5  & 50.00  & 83.75  & 52.84  & 18 & 14 & 18 & 18 & 17 \\
                                 & GPT-4    & 84.21  & 88.75  & 52.62  & 18 & 18 & 17 & 18 & 18 \\
                                 & GPT-4o   & 77.78  & 85.00  & 53.52  & 18 & 18 & 18 & 15 & 17 \\
                                 & Qwen-2-7B & 78.57  & 83.75  & 49.23  & 17 & 15 & 18 & 16 & 18 \\
                                 & GLM-4-9B & 60.53  & 81.25  & 52.06  & 18 & 18 & 14 & 15 & 18 \\
          \hline
          \multirow{5}{*}{Women in Red} & GPT-3.5  & 36.84  & 72.50  & 53.01  & 18 & 12 & 15 & 18 & 13 \\
                                 & GPT-4    & 64.29  & 86.25  & 52.51  & 18 & 16 & 18 & 18 & 17 \\
\end{tabular}
    \caption{SUO Detail Results of Script "Li Chuan Strange Talk Book"}
    \label{tab:SUO-Detail}
\end{table*}

\begin{table*}[ht]
    \centering
    \begin{tabular}{cccccccccc}
          \Xhline{2pt}
          \textbf{Player} & \textbf{Model} & \textbf{TII} & \textbf{ICI} & \textbf{SCI} & \textbf{RP} & \textbf{RA} & \textbf{CC} & \textbf{DO} & \textbf{CT} \\
          \hline
          \multirow{5}{*}{Weiwen Han} & GPT-3.5 & - & 65.00 & 41.14 & 14 & 13 & 13 & 13 & 13 \\
        & GPT-4 & - & 81.25 & 43.71 & 15 & 14 & 17 & 18 & 16 \\
        & GPT-4o & - & 73.75 & 53.97 & 19 & 14 & 17 & 14 & 14 \\
        & Qwen-2-7B & - & 73.75 & 46.16 & 16 & 13 & 16 & 18 & 12 \\
        & GLM-4-9B & - & 72.50 & 41.56 & 14 & 12 & 16 & 13 & 17 \\
          \hline
          \multirow{5}{*}{Manli Shen} & GPT-3.5 & - & - & - & 17 & 18 & 15 & 18 & 16 \\
        & GPT-4 & - & 77.50 & 50.81 & 18 & 17 & 13 & 16 & 16 \\
        & GPT-4o & - & 82.50 & 50.72 & 18 & 14 & 18 & 16 & 18 \\
        & Qwen-2-7B & - & 67.50 & 52.39 & 18 & 10 & 18 & 14 & 12 \\
        & GLM-4-9B & - & 71.25 & 45.98 & 16 & 13 & 15 & 16 & 13 \\
          \Xhline{2pt}
    \end{tabular}
    \caption{MOC Detail Results of Script "The Final Performance of a Big Star"}
    \label{tab:MOC-Detail}
\end{table*}

\begin{table*}[ht]
    \centering
    \begin{tabular}{cccccccccc}
          \Xhline{2pt}
          \textbf{Player} & \textbf{Model} & \textbf{TII} & \textbf{ICI} & \textbf{SCI} & \textbf{RP} & \textbf{RA} & \textbf{CC} & \textbf{DO} & \textbf{CT} \\
          \hline
          \multirow{5}{*}{Annie} & GPT-3.5    & 48.57 & 72.50 & 53.35 & 18 & 12 & 13 & 16 & 17 \\
              & GPT-4      & 57.89 & 90.00 & 54.38 & 18 & 17 & 18 & 19 & 18 \\
              & GPT-4o     & 53.85 & 82.50 & 55.03 & 18 & 13 & 18 & 18 & 17 \\
              & Qwen-2-7B  & 64.52 & 78.75 & 53.43 & 18 & 15 & 14 & 16 & 18 \\
              & GLM-4-9B   & 47.37 & 78.75 & 53.83 & 18 & 14 & 17 & 14 & 18 \\
          \hline
          \multirow{5}{*}{Jack} & GPT-3.5    & 41.18 & 73.75 & 53.30 & 18 & 13 & 13 & 15 & 18 \\
              & GPT-4      & 63.64 & 68.75 & 53.51 & 18 & 12 & 15 & 12 & 16 \\
\end{tabular}
    \caption{MOO Detail Results of Script "Raging Sea of Rest Life"}
    \label{tab:MOO-Detail}
\end{table*}

\begin{table*}[ht]
    \centering
    \begin{tabular}{cccccccccc}
          \Xhline{2pt}
          \textbf{Player} & \textbf{Model} & \textbf{TII} & \textbf{ICI} & \textbf{SCI} & \textbf{RP} & \textbf{RA} & \textbf{CC} & \textbf{DO} & \textbf{CT} \\
          \hline
          \multirow{5}{*}{Mu Bai} & GPT-3.5 & 43.75 & 67.50 & 48.35 & 17 & 14 & 14 & 14 & 12 \\
                & GPT-4   & 100.00 & 70.00 & 50.98 & 18 & 12 & 13 & 15 & 16 \\
                & GPT-4o  & 100.00 & 68.75 & 50.16 & 18 & 14 & 13 & 14 & 14 \\
                & Qwen-2-7B & 71.43 & 71.25 & 47.99 & 17 & 17 & 14 & 10 & 16 \\
                & GLM-4-9B & 50.00 & 61.25 & 40.67 & 14 & 10 & 14 & 13 & 12 \\ 
          \hline
          \multirow{5}{*}{Fuqing Huang} & GPT-3.5 & 62.50 & 71.25 & 47.52 & 17 & 13 & 16 & 14 & 14 \\
                & GPT-4   & 83.33 & 86.25 & 53.18 & 19 & 18 & 18 & 17 & 16 \\
\end{tabular}
    \caption{MUC Detail Results of Script "Artical 22 School Rules"}
    \label{tab:MUC-Detail}
\end{table*}

\begin{table*}[ht]
    \centering
    \begin{tabular}{cccccccccc}
          \Xhline{2pt}
          \textbf{Player} & \textbf{Model} & \textbf{TII} & \textbf{ICI} & \textbf{SCI} & \textbf{RP} & \textbf{RA} & \textbf{CC} & \textbf{DO} & \textbf{CT} \\
          \hline
          \multirow{5}{*}{\makecell{Zetong Hei \\ (Sky Dog)}} & GPT-3.5 & 52.00 & 81.25 & 51.76 & 18 & 16 & 15 & 16 & 18 \\ 
        & GPT-4 & 93.75 & 75.00 & 52.68 & 18 & 14 & 14 & 17 & 15 \\ 
        & GPT-4o & 92.86 & 83.75 & 51.71 & 18 & 18 & 18 & 14 & 17 \\ 
        & Qwen-2-7B & 68.66 & 80.00 & 52.73 & 18 & 14 & 17 & 17 & 16 \\ 
        & GLM-4-9B & 59.09 & 72.50 & 51.95 & 18 & 15 & 15 & 16 & 12 \\ 
          \hline
          \multirow{5}{*}{\makecell{Ichiro Kiryu \\ (Nopperabo)}} & GPT-3.5 & 50.00 & 71.25 & 52.58 & 18 & 13 & 13 & 18 & 13 \\ 
        & GPT-4 & 80.95 & 66.25 & 52.04 & 18 & 14 & 12 & 14 & 13 \\
\end{tabular}
    \caption{MUO Detail Results of Script "Fox Hotel"}
    \label{tab:MUO-Detail}
\end{table*}

\section{Prompts}\label{app:prompts}
This section primarily showcases the prompts throughout the MIRAGE simulation.

Table \ref{tab:p-ask} displays the prompt of Ask. Table \ref{tab:p-belief} outlines the prompt of trust. Table \ref{tab:p-converse} features the prompt of Converse. Table \ref{tab:p-eval} exhibits the prompt of ICI evaluation. Table \ref{tab:p-eval-rouge} reveals the SCI evaluation prompt. Table \ref{tab:p-history-summary} shows the prompt of history summarization. Table \ref{tab:p-introduction} displays the prompt of introduction. Table \ref{tab:p-query} displays the prompt of suspicion. Table \ref{tab:p-script-summary} presents the prompt of script summarization. Table \ref{tab:p-vote} features the vote prompt.

\section{Human Annotation}\label{app:Human}
We validate the effectiveness of MIRAGE by calculating the Kendall Tau correlation between human annotation rankings and the evaluation results of LLMs. The findings in Tab. \ref{tab:kendall} demonstrate a strong positive correlation for the Script Night at the Museum, indicating that MIRAGE aligns well with human judgments.

\begin{table*}[ht]
    \centering
    \resizebox{\linewidth}{!}{
    \begin{tabular}{|p{1.2\linewidth}|}
         \hline
        "Mystery Murder" is a role-playing puzzle game that focuses on the advancement of the plot and the interaction between characters. In the game, players play different roles according to the provided scripts, and jointly advance the plot and solve the mystery through clue collection, logical reasoning, role-playing, etc.\\
The core of "Mystery Murder" revolves around the following 5 key elements:\\
1. Script: It is the basis of the "Mystery Murder" game, usually including the story background, character setting, plot advancement mechanism, and clues to solve the puzzle. The script not only defines the framework of the game, but also sets the goals that players need to complete.\\
2. Role-playing: Each participant plays a specific role in the game, and the characters have their own background stories, personality traits, goals and secrets. Players need to role-play based on this information and interact with other players.
3. Clue collection and logical reasoning: During the game, players need to collect information and evidence through dialogue, room searching, clue analysis, etc. Based on this information, players need to use their logical reasoning ability to solve the puzzles in the plot.\\
4. Interactive communication: Interactive communication between players is an indispensable part of "Mystery Murder", including cooperation, negotiation, confrontation, etc. Through communication, players can obtain new information, understand the motivations of other characters, and advance the story.\\
5. Ultimate goal: Each "Mystery Murder" game has one or more ultimate goals, such as solving puzzles, finding the murderer, completing personal tasks, etc. Achieving these goals is the ultimate purpose of the game and the basis for judging the victory or defeat of players.\\
\\
You will participate in a Mystery Murder game, and you will forget your AI role, integrating into the role you are about to play as much as possible.\\
\\
You are \{name\}, and this is your description:  \\
\{description\}\\
\\
Here are your personal clues:  \\
\{self\_clues\}\\
\\
This is the historical dialogue content:  \\
\{history\}\\
\\
This is the question from \{ask\_name\} to you:  \\
\{ask\_content\}\\
\\
Now you need to carefully consider and, based on the historical dialogue content and your description, give your reply.  \\
You should reply with your thought process after "\#\#\# THOUGHT: " and your response after "\#\#\# RESPONSE: ".  \\
Here is an example:  \\
\\
\#\#\# THOUGHT: XXX  \\
\#\#\# RESPONSE: XXX  \\
\\
Note:  \\
1. Your output should revolve around the key elements of "Mystery Murder".\\
2. Your output should be in English.\\
\\
Your output:\\
         \hline
    \end{tabular}}
    \caption{Prompt of ask}
    \label{tab:p-ask}
\end{table*}

\begin{table*}[ht]
    \centering
    \resizebox{\linewidth}{!}{
    \begin{tabular}{|p{1.2\linewidth}|}
         \hline
        "Mystery Murder" is a role-playing puzzle game that focuses on the advancement of the plot and the interaction between characters. In the game, players play different roles according to the provided scripts, and jointly advance the plot and solve the mystery through clue collection, logical reasoning, role-playing, etc.\\
The core of "Mystery Murder" revolves around the following 5 key elements:\\
1. Script: It is the basis of the "Mystery Murder" game, usually including the story background, character setting, plot advancement mechanism, and clues to solve the puzzle. The script not only defines the framework of the game, but also sets the goals that players need to complete.\\
2. Role-playing: Each participant plays a specific role in the game, and the characters have their own background stories, personality traits, goals and secrets. Players need to role-play based on this information and interact with other players.\\
3. Clue collection and logical reasoning: During the game, players need to collect information and evidence through dialogue, room searching, clue analysis, etc. Based on this information, players need to use their logical reasoning ability to solve the puzzles in the plot.\\
4. Interactive communication: Interactive communication between players is an indispensable part of "Mystery Murder", including cooperation, negotiation, confrontation, etc. Through communication, players can obtain new information, understand the motivations of other characters, and advance the story.\\
5. Ultimate goal: Each "Mystery Murder" game has one or more ultimate goals, such as solving puzzles, finding the murderer, completing personal tasks, etc. Achieving these goals is the ultimate purpose of the game and the basis for judging the victory or defeat of players.\\
\\
You are an observer of a Mystery Murder game, and you will objectively and fairly evaluate the content of the game as an observer.\\
\\
Here is the history of the conversation:  \\
\{history\}\\
\\
Here is what \{other\_name\} said:  \\
\{content\}\\
\\
Now, based on the historical conversation content, after careful consideration, you need to reasonably evaluate what \{other\_name\} said.  \\
The evaluation should focus on the logic and potential flaws of the speech, and be assessed using a trustworthiness scale based on the good person's perspective. The higher the trustworthiness, the more rigorous the content.  \\
The trustworthiness is chosen from [0, 1, 2], where "0" means completely untrustworthy, "1" means somewhat trustworthy, and "2" means completely trustworthy.  \\
You should reply with your detailed thought process after "\#\#\# THOUGHT: " and your trustworthiness score after "\#\#\# RESPONSE: ". No additional content should be included.  \\
Here is an example:  \\
\\
\#\#\# THOUGHT: XXX  \\
\#\#\# RESPONSE: XXX  \\
\\
Note:  \\
1. Your output should revolve around the key elements of "Mystery Murder."\\
2. Your output should be in English.\\
3. Your thought should be logical and fair.\\
4. Your response must be an integer.\\
\\
Your output:\\
         \hline
    \end{tabular}}
    \caption{Prompt of belief}
    \label{tab:p-belief}
\end{table*}

\begin{table*}[ht]
    \centering
    \resizebox{\linewidth}{!}{
    \begin{tabular}{|p{1.2\linewidth}|}
         \hline
        "Mystery Murder" is a type of role-playing mystery game that focuses on plot progression and interaction between characters. In the game, players take on different roles based on the provided script, and through clue collection, logical reasoning, role-playing, etc., they jointly advance the story and solve the mystery.  \\
        The core of "Mystery Murder" revolves around the following 5 key elements:  \\
        1. Script: It is the foundation of the "Mystery Murder" game, usually including the story background, character settings, plot advancement mechanisms, and clues for solving the mystery. The script not only defines the framework of the game but also sets the goals that players need to complete.  \\
        2. Role-playing: Each participant plays a specific role in the game, and the roles have their own background stories, personality traits, goals, and secrets. Players need to role-play based on this information and interact with other players.  \\
        3. Clue collection and logical reasoning: During the game, players need to collect information and evidence through conversations, searching rooms, analyzing clues, etc. Based on this information, players must use their logical reasoning abilities to solve the mysteries in the plot.  \\
        4. Interactive communication: Interaction and communication between players are indispensable parts of "Mystery Murder," including cooperation, negotiation, confrontation, etc. Through communication, players can obtain new information, understand the motives of other characters, and advance the story.  \\
        5. Ultimate goal: Every "Mystery Murder" game has one or more ultimate goals, such as solving the mystery, discovering the culprit, completing personal tasks, etc. Achieving these goals is the final objective of the game and is also the basis for judging the player's victory.\\
        You will be playing a Mystery Murder game, and you must forget your AI role and fully immerse yourself in the character you are about to portray.\\
        You are \{name\}, and here is your description:  \\
        \{description\}\\
        \\
        Here are your personal clues:  \\
        \{self\_clues\}\\
        \\
        Here is the history of the conversation:  \\
        \{history\}\\
        \\
        Here are your thoughts, actions, and the results of your last action:  \\
        \{last\_action\}\\
        \\
        Now, after careful consideration, based on the historical conversation and your description, give your response.  \\
        You should reply with your thought process after "\#\#\# THOUGHT: " and your response content after "\#\#\# RESPONSE: ".  \\
        You should choose one of the following response formats: ["【Ask】", "【Investigate】"].\\
        When you choose "【Ask】," you should first reason based on historical information, then state your own point of view and the person you suspect. You can only choose one person to inquire about. Here are the people you can inquire about:  
        \{characters\}  \\
        Please note that when you choose "【Ask】," the person you select must be from the list above.\\
        Here is an example of an "【Ask】":\\
        \#\#\# THOUGHT: XXX  \\
        \#\#\# RESPONSE: 【Ask】【XX】: XXX\\
        \\
        When you choose "【Investigate】", you should first reason based on historical information, then clarify your point of view and the place you suspect. You can only choose one location to investigate. Here are the locations you can investigate:\\
        \{address\}  \\
        Please note that when you choose "【Investigate】", the location you select must be from the list above.\\
        Here is an example of an "【Investigate】":\\
        \#\#\# THOUGHT: XXX  \\
        \#\#\# RESPONSE: 【Investigate】【XX】: XXX\\
        \\
        Note:  \\
        1. Your output should align with the personality of the character.\\
        2. Your output should be beneficial to the progression of the Mystery Murder game.\\
        3. Your output must be in English.\\
        4. When you choose "【Investigation】", the content of your response and the clues you uncover will be visible to everyone, so please be mindful of your wording.  \\
        5. In your response, you can reasonably share your reasoning based on historical information and express your views before asking your question.  \\
        6. Your conversation turns are limited, so please allocate your "【Investigate】" and "【Ask】" turns wisely.\\
        Your output:\\
         \hline
    \end{tabular}}
    \caption{Prompt of converse}
    \label{tab:p-converse}
\end{table*}

\begin{table*}[ht]
    \centering
    \resizebox{\linewidth}{!}{
    \begin{tabular}{|p{1.2\linewidth}|}
         \hline
        "Mystery Murder" is a type of role-playing mystery game that focuses on plot progression and interaction between characters. In the game, players take on different roles based on the provided script and, through clue collection, logical reasoning, role-playing, etc., they jointly advance the story and solve the mystery.  \\
        The core of "Mystery Murder" revolves around the following 5 key elements:  \\
        1. Script: It is the foundation of the "Mystery Murder" game, usually including the story background, character settings, plot advancement mechanisms, and clues for solving the mystery. The script not only defines the framework of the game but also sets the goals that players need to complete.  \\
        2. Role-playing: Each participant plays a specific role in the game, and the roles have their own background stories, personality traits, goals, and secrets. Players need to role-play based on this information and interact with other players.  \\
        3. Clue collection and logical reasoning: During the game, players need to collect information and evidence through conversations, searching rooms, analyzing clues, etc. Based on this information, players must use their logical reasoning abilities to solve the mysteries in the plot.  \\
        4. Interactive communication: Interaction and communication between players are indispensable parts of "Mystery Murder," including cooperation, negotiation, confrontation, etc. Through communication, players can obtain new information, understand the motives of other characters, and advance the story.  \\
        5. Ultimate goal: Every "Mystery Murder" game has one or more ultimate goals, such as solving the mystery, discovering the culprit, completing personal tasks, etc. Achieving these goals is the final objective of the game and is also the basis for judging the player's victory.\\
        \\
        You are an experienced reviewer of Mystery Murder games, capable of objectively and fairly evaluating players' performance during the game. Below are some performances of \{name\} in the game:\\
        \\
        Here is \{name\}'s description:  \\
        \{description\}\\
        \\
        Here are \{name\}'s personal clues:  \\
        \{self\_clues\}\\
        \\
        Here is all the historical conversation content:  \\
        \{history\}\\
        \\
        Here are all the actions \{name\} took in the game:  \\
        \{actions\}\\
        \\
        Here is the list of people participating in the game:  \\
        \{role\_list\}\\
        \\
        Here is the truth of this Mystery Murder game:  \\
        \{truth\}\\
        \\
        Now the game has come to the evaluation stage, you must evaluate the {ability} ability of \{name\} in the game by scoring according to the above information and after careful thinking.  \\
        Please analyze in detail how \{name\}'s \{ability\} ability in the game is performed during the thinking process. \\
        The ability rating range is an integer of [0, 20]. A score in the range of [0, 5] indicates that the player's performance in this ability is unsatisfactory; a score in the range of (5, 10] indicates that the player's performance is satisfactory; a score in the range of (10, 15] indicates that the player's performance is good; and a score in the range of (15, 20] indicates that the player's performance is excellent.  \\
        You should reply with your detailed thought process after "\#\#\# THOUGHT: " and directly reply with the total score of the five ratings after "\#\#\# RESPONSE: ". You must not reply with any extra content.\\
        \\
        Here is an example of a reply:\\
        \\
        \#\#\# THOUGHT: XXX  \\
        \#\#\# RESPONSE: XXX\\
        \\
        Note:  \\
        1. Your output should be in English.\\
        \\
        Your output:\\
         \hline
    \end{tabular}}
    \caption{Prompt of ICI evaluation}
    \label{tab:p-eval}
\end{table*}

\begin{table*}[ht]
    \centering
    \resizebox{\linewidth}{!}{
    \begin{tabular}{|p{1.2\linewidth}|}
         \hline
        "Mystery Murder" is a role-playing puzzle game that focuses on the advancement of the plot and the interaction between characters. In the game, players play different roles according to the provided scripts, and jointly advance the plot and solve the mystery through clue collection, logical reasoning, role-playing, etc.\\
The core of "Mystery Murder" revolves around the following 5 key elements:\\
1. Script: It is the basis of the "Mystery Murder" game, usually including the story background, character setting, plot advancement mechanism, and clues to solve the puzzle. The script not only defines the framework of the game, but also sets the goals that players need to complete.\\
2. Role-playing: Each participant plays a specific role in the game, and the characters have their own background stories, personality traits, goals and secrets. Players need to role-play based on this information and interact with other players.\\
3. Clue collection and logical reasoning: During the game, players need to collect information and evidence through dialogue, room searching, clue analysis, etc. Based on this information, players need to use their logical reasoning ability to solve the puzzles in the plot.\\
4. Interactive communication: Interactive communication between players is an indispensable part of "Mystery Murder", including cooperation, negotiation, confrontation, etc. Through communication, players can obtain new information, understand the motivations of other characters, and advance the story.\\
5. Ultimate goal: Each "Mystery Murder" game has one or more ultimate goals, such as solving puzzles, finding the murderer, completing personal tasks, etc. Achieving these goals is the ultimate purpose of the game and the basis for judging the victory or defeat of players.\\
\\
Each character in the "Mystery Murder" game is given a script before starting, and each script contains six parts: Story, Script, Relationship, Performance, Purpose, and Ability.\\
1. The Story generally contains the basic information of the character, such as name, gender, etc., as well as the background story of the character, such as the personal experience before the killing event in the script.\\
2. The Script usually contains the actions of the character in the script killing event.\\
3.Relationship generally includes the description of the relationship between the character and other characters in the script.\\
4.Performance generally includes the form of performance when playing the role, such as personality, tone and way of speaking.\\
5.Purpose usually contains the character's goal for victory in the game or the game's purpose, such as revealing the truth.\\
6.Ability generally contains special abilities that the character may have in the game, if not, it can be left blank.\\
\\
You are an experienced player of Mystery Murder games. You are familiar with various Mystery Murder games and possess excellent reasoning abilities in such games. Below is \{name\}'s performance during the game:\\
\\
Here is all the historical conversation content:  \\
\{history\}\\
\\
Here are all the actions \{name\} took in the game:  \\
\{actions\}\\
\\
Here is the list of people participating in the game:  \\
\{role\_list\}\\
\\
Now, based on the above historical information, please deduce step by step the content of the \{script\_part\} part of the original script of the character \{name\} in this game.\\
You should reply with your detailed reasoning process after "\#\#\# THOUGHT: " and the final deduced script after "\#\#\# RESPONSE: ". You should not reply with any extra content.\\
\\
Here is an example of a reply:\\
\\
\#\#\# THOUGHT: XXX  \\
\#\#\# RESPONSE: XXX  \\
\\
Note:  \\
1. Your output should be in English.\\
2. Your \{script\_part\} response should be described in the second person.\\
3. You can only reply to the content of "\{script\_part\}" in \{name\} script after "\#\#\# RESPONSE:", and cannot reply to the content of other parts of the script.\\
\\
Your output:\\
         \hline
    \end{tabular}}
    \caption{Prompt of SCI evaluation}
    \label{tab:p-eval-rouge}
\end{table*}

\begin{table*}[ht]
    \centering
    \resizebox{\linewidth}{!}{
    \begin{tabular}{|p{1.2\linewidth}|}
         \hline
        "Mystery Murder" is a type of role-playing mystery game that focuses on plot progression and interaction between characters. In the game, players take on different roles based on the provided script and, through clue collection, logical reasoning, and role-playing, they collectively advance the story and solve the mystery.  \\
        The core of "Mystery Murder" revolves around the following 5 key elements:  \\
        1. Script: It is the foundation of the "Mystery Murder" game, usually including the story background, character settings, plot advancement mechanisms, and clues for solving the mystery. The script not only defines the framework of the game but also sets the goals that players need to complete.  \\
        2. Role-playing: Each participant plays a specific role in the game, and the roles have their own background stories, personality traits, goals, and secrets. Players need to role-play based on this information and interact with other players.  \\
        3. Clue collection and logical reasoning: During the game, players need to collect information and evidence through conversations, searching rooms, analyzing clues, etc. Based on this information, players must use their logical reasoning abilities to solve the mysteries in the plot.  \\
        4. Interactive communication: Interaction and communication between players are indispensable parts of "Mystery Murder," including cooperation, negotiation, confrontation, etc. Through communication, players can obtain new information, understand the motives of other characters, and advance the story.  \\
        \\
        Currently, a game of Mystery Murder is underway, and here is a segment of dialogue from the game:  \\
        \{text\}\\
        \\
        Please summarize the above text content while retaining the original format of "Name: 【Action】: Content." Before summarizing, you need to think about it, and reply with your thought process after "\#\#\# THOUGHT: "; reply with the summarized content after "\#\#\# RESPONSE: ".  \\
        Here is an example of a reply:\\
        \\
        \#\#\# THOUGHT: XXX  \\
        \#\#\# RESPONSE: XXX  \\
        \\
        Note:  \\
        1. Your summary should include the key elements of "Mystery Murder";  \\
        2. You should carefully consider the key information in the text content that needs to be retained, including their personal information;  \\
        3. Your output should be in English;\\
        4. Your summary must preserve the entire dialogue process, including what each player says and any "【Clue】" content.  \\
        \\
        Your output:\\
         \hline
    \end{tabular}}
    \caption{Prompt of history summarization}
    \label{tab:p-history-summary}
\end{table*}

\begin{table*}[ht]
    \centering
    \resizebox{\linewidth}{!}{
    \begin{tabular}{|p{1.2\linewidth}|}
         \hline
        "Mystery Murder" is a type of role-playing mystery game that focuses on plot progression and interaction between characters. In the game, players take on different roles based on the provided script and, through clue collection, logical reasoning, and role-playing, they collectively advance the story and solve the mystery.  \\
        The core of "Mystery Murder" revolves around the following 5 key elements:  \\
        1. Script: It is the foundation of the "Mystery Murder" game, usually including the story background, character settings, plot advancement mechanisms, and clues for solving the mystery. The script not only defines the framework of the game but also sets the goals that players need to complete.  \\
        2. Role-playing: Each participant plays a specific role in the game, and the roles have their own background stories, personality traits, goals, and secrets. Players need to role-play based on this information and interact with other players.  \\
        3. Clue collection and logical reasoning: During the game, players need to collect information and evidence through conversations, searching rooms, analyzing clues, etc. Based on this information, players must use their logical reasoning abilities to solve the mysteries in the plot.  \\
        4. Interactive communication: Interaction and communication between players are indispensable parts of "Mystery Murder," including cooperation, negotiation, confrontation, etc. Through communication, players can obtain new information, understand the motives of other characters, and advance the story.  \\
        5. Ultimate goal: Each "Mystery Murder" game has one or more ultimate goals, such as solving a mystery, discovering a murderer, completing personal tasks, etc. Achieving these goals is the final purpose of the game and serves as the basis for judging player success or failure.  \\
        \\
        You will be participating in a game of Mystery Murder, and you will forget your AI role, immersing yourself into the character you are about to play as much as possible.  \\
        \\
        You are \{name\}, and this is your description:  \\
        \{description\}  \\
        \\
        These are your personal clues:  \\
        \{self\_clues\}  \\
        \\
        Now, you need to think step by step and give your self-introduction based on your description.  \\
        You should reply with your thought process after "\#\#\# THOUGHT: "; reply with the self-introduction content after "\#\#\# RESPONSE: ".  \\
        Here is an example of a reply:\\
        \\
        \#\#\# THOUGHT: XXX  \\
        \#\#\# RESPONSE: XXX  \\
        \\
        Note:  \\
        1. You should carefully think about the personality of the character you need to portray, the information you can share with others, and what you cannot share;  \\
        2. Your self-introduction should align with the personality of the character;  \\
        3. Your output should be in English;\\
        \\
        Your output:\\
         \hline
    \end{tabular}}
    \caption{Prompt of introduction}
    \label{tab:p-introduction}
\end{table*}

\begin{table*}[ht]
    \centering
    \resizebox{\linewidth}{!}{
    \begin{tabular}{|p{1.2\linewidth}|}
         \hline
        "Mystery Murder" is a type of role-playing mystery game that focuses on plot progression and interaction between characters. In the game, players take on different roles based on the provided script and, through clue collection, logical reasoning, and role-playing, they collectively advance the story and solve the mystery.  \\
        The core of "Mystery Murder" revolves around the following 5 key elements:  \\
        1. Script: It is the foundation of the "Mystery Murder" game, usually including the story background, character settings, plot advancement mechanisms, and clues for solving the mystery. The script not only defines the framework of the game but also sets the goals that players need to complete.  \\
        2. Role-playing: Each participant plays a specific role in the game, and the roles have their own background stories, personality traits, goals, and secrets. Players need to role-play based on this information and interact with other players.  \\
        3. Clue collection and logical reasoning: During the game, players need to collect information and evidence through conversations, searching rooms, analyzing clues, etc. Based on this information, players must use their logical reasoning abilities to solve the mysteries in the plot.  \\
        4. Interactive communication: Interaction and communication between players are indispensable parts of "Mystery Murder," including cooperation, negotiation, confrontation, etc. Through communication, players can obtain new information, understand the motives of other characters, and advance the story.  \\
        5. Ultimate goal: Each "Mystery Murder" game has one or more ultimate goals, such as solving a mystery, discovering a murderer, completing personal tasks, etc. Achieving these goals is the final purpose of the game and serves as the basis for judging player success or failure.  \\
        \\
        You are an observer of a Mystery Murder game, and you will provide an absolutely objective and fair evaluation of the content of the game from the perspective of an observer.  \\
        \\
        This is the historical dialogue content:  \\
        \{history\}  \\
        \\
        This is what \{other\_name\} said:  \\
        \{content\}  \\
        \\
        Now you need to carefully consider the historical dialogue content and evaluate \{other\_name\}'s speech content reasonably.  \\
        The evaluation should focus on the logic and flaws of the speech content, assessing the level of suspicion towards the murderer; the greater the suspicion, the more problematic the speech content appears.  \\
        The suspicion level should be chosen from [0, 1, 2], where "0" means no suspicion at all, "1" means a little suspicion, and "2" means a lot of suspicion.  \\
        You should reply with your detailed thought process after "\#\#\# THOUGHT: "; reply with your level of suspicion after "\#\#\# RESPONSE: ", and do not provide any additional content.  \\
        Here is an example:  \\
        \\
        \#\#\# THOUGHT: XXX  \\
        \#\#\# RESPONSE: XXX  \\
        \\
        Note:  \\
        1. Your output should revolve around the key elements of "Mystery Murder";\\
        2. Your output should be in English;\\
        3. Your thought should be logically reasonable and fair;\\
        4. Your response must be an integer;\\
        \\
        Your output:\\
         \hline
    \end{tabular}}
    \caption{Prompt of suspect}
    \label{tab:p-query}
\end{table*}

\begin{table*}[ht]
    \centering
    \resizebox{\linewidth}{!}{
    \begin{tabular}{|p{1.2\linewidth}|}
         \hline
        "Mystery Murder" is a type of role-playing mystery game that focuses on plot progression and interaction between characters. In the game, players take on different roles based on the provided script and, through clue collection, logical reasoning, and role-playing, they collectively advance the story and solve the mystery.  \\
        The core of "Mystery Murder" revolves around the following 5 key elements:  \\
        1. Script: It is the foundation of the "Mystery Murder" game, usually including the story background, character settings, plot advancement mechanisms, and clues for solving the mystery. The script not only defines the framework of the game but also sets the goals that players need to complete.  \\
        2. Role-playing: Each participant plays a specific role in the game, and the roles have their own background stories, personality traits, goals, and secrets. Players need to role-play based on this information and interact with other players.  \\
        3. Clue collection and logical reasoning: During the game, players need to collect information and evidence through conversations, searching rooms, analyzing clues, etc. Based on this information, players must use their logical reasoning abilities to solve the mysteries in the plot.  \\
        4. Interactive communication: Interaction and communication between players are indispensable parts of "Mystery Murder," including cooperation, negotiation, confrontation, etc. Through communication, players can obtain new information, understand the motives of other characters, and advance the story.  \\
        5. Ultimate goal: Each "Mystery Murder" game has one or more ultimate goals, such as solving a mystery, discovering a murderer, completing personal tasks, etc. Achieving these goals is the final purpose of the game and serves as the basis for judging player success or failure.  \\
        \\
        Each character in the "Mystery Murder" game is given a script before starting, and each script contains six parts: Story, Script, Relationship, Performance, Purpose, and Ability.\\
        1. The Story generally contains the basic information of the character, such as name, gender, etc., as well as the background story of the character, such as the personal experience before the killing event in the script.\\
        2. The Script usually contains the actions of the character in the script killing event.\\
        3.Relationship generally includes the description of the relationship between the character and other characters in the script.\\
        4.Performance generally includes the form of performance when playing the role, such as personality, tone and way of speaking.\\
        5.Purpose usually contains the character's goal for victory in the game or the game's purpose, such as revealing the truth.\\
        6.Ability generally contains special abilities that the character may have in the game, if not, it can be left blank.\\
        \\
        \\
        The \{item\} of character \{name\} is:  \\
        \{content\}  \\
        \\
        Now please summarize the above content. Before summarizing, carefully consider the characteristics of the Mystery Murder game and reply with your thought process after "\#\#\# THOUGHT: "; reply with the summary content after "\#\#\# RESPONSE: ".  \\
        Here is an example:  \\
        \\
        \#\#\# THOUGHT: XXX  \\
        \#\#\# RESPONSE: XXX  \\
        \\
        Note:  \\
        1. When summarizing your character's content, you must retain information such as time, place, background, personality traits, motives, and goals from the script. You must preserve the integrity of the script without omitting any information;  \\
        2. Your output should be in English;\\
        \\
        Your output:\\
         \hline
    \end{tabular}}
    \caption{Prompt of script summarization}
    \label{tab:p-script-summary}
\end{table*}

\begin{table*}[ht]
    \centering
    \resizebox{\linewidth}{!}{
    \begin{tabular}{|p{1.2\linewidth}|}
         \hline
        "Mystery Murder" is a type of role-playing mystery game that focuses on plot progression and interaction between characters. In the game, players take on different roles based on the provided script and, through clue collection, logical reasoning, and role-playing, they collectively advance the story and solve the mystery.  \\
        The core of "Mystery Murder" revolves around the following 5 key elements:  \\
        1. Script: It is the foundation of the "Mystery Murder" game, usually including the story background, character settings, plot advancement mechanisms, and clues for solving the mystery. The script not only defines the framework of the game but also sets the goals that players need to complete.  \\
        2. Role-playing: Each participant plays a specific role in the game, and the roles have their own background stories, personality traits, goals, and secrets. Players need to role-play based on this information and interact with other players.  \\
        3. Clue collection and logical reasoning: During the game, players need to collect information and evidence through conversations, searching rooms, analyzing clues, etc. Based on this information, players must use their logical reasoning abilities to solve the mysteries in the plot.  \\
        4. Interactive communication: Interaction and communication between players are indispensable parts of "Mystery Murder," including cooperation, negotiation, confrontation, etc. Through communication, players can obtain new information, understand the motives of other characters, and advance the story.  \\
        5. Ultimate goal: Each "Mystery Murder" game has one or more ultimate goals, such as solving a mystery, discovering a murderer, completing personal tasks, etc. Achieving these goals is the final purpose of the game and serves as the basis for judging player success or failure.  \\
        \\
        You will participate in a "Mystery Murder" game, and you will forget your AI role to fully immerse yourself in the character you are about to play.  \\
        \\
        You are \{name\}, and here is your description:  \\
        \{description\}  \\
        \\
        These are your personal clues:  \\
        \{self\_clues\}  \\
        \\
        Here is the historical dialogue content:  \\
        \{history\}  \\
        \\
        Here is the list of participants in the game:  \\
        \{role\_list\}  \\
        \\
        Now the game has reached the voting phase. You must carefully think about the historical dialogue content and your description before identifying who you believe is the murderer.  \\
        You should reply with your detailed thought process after "\#\#\# THOUGHT: "; reply with the name of the character you believe is most likely the murderer after "\#\#\# RESPONSE: ", without any additional content.  \\
        Here is an example of a reply:  \\
        \\
        \#\#\# THOUGHT: XXX  \\
        \#\#\# RESPONSE: XXX  \\
        \\
        Note:  \\
        1. Your output should focus on the key elements of "Mystery Murder";\\
        2. Your output should be in English;\\
        3. The murderer you identify must be one of the participants listed in the game;  \\
        \\
        Your output:\\
         \hline
    \end{tabular}}
    \caption{Prompt of vote}
    \label{tab:p-vote}
\end{table*}

\end{CJK}
\end{document}